\documentclass[pdflatex,sn-mathphys-num]{sn-jnl}

\usepackage{graphicx}%
\usepackage{multirow}%
\usepackage{amsmath,amssymb,amsfonts}%
\usepackage{amsthm}%
\usepackage{mathrsfs}%
\usepackage[title]{appendix}%
\usepackage{xcolor}%
\usepackage{textcomp}%
\usepackage{manyfoot}%
\usepackage{booktabs}%
\usepackage{algorithm}%
\usepackage{algorithmicx}%
\usepackage{algpseudocode}%
\usepackage{listings}%
\usepackage{tabularx}
\usepackage{float} 
\usepackage{tcolorbox}
\tcbuselibrary{breakable,skins}
\definecolor{lightestgray}{gray}{0.95}
\definecolor{darkgray}{gray}{0.30}
\lstdefinelanguage{json}{
  showstringspaces=false,
  breaklines=true,
  breakatwhitespace=false,
  columns=fullflexible,
  keepspaces=true,
  tabsize=2,
  morestring=[b]",
  morecomment=[l]{//},
  morecomment=[s]{/*}{*/},
  postbreak=\mbox{\textcolor{gray}{$\hookrightarrow$}\space}
}
\lstdefinestyle{jsonbox}{
  language=json,
  basicstyle=\ttfamily\scriptsize,
  numberstyle=\scriptsize,
  frame=none,
  numbers=left,
  numbersep=2pt,
  xleftmargin=2.5ex,
  framexleftmargin=0pt,
  linewidth=\linewidth,
  aboveskip=0.2\baselineskip,
  belowskip=0.2\baselineskip
}

\theoremstyle{thmstyleone}%
\theoremstyle{thmstyletwo}%

\theoremstyle{thmstylethree}%

\newcommand{\llm}{\textsc{LLM}}
\newcommand{\sdsexample}{SULFURIC ACID 1-51\%)}

\begin{document}

\title[Article Title]{Benchmarking Large Language Models for Safety Data Extraction}


\author[1]{\fnm{Jonas} \sur{Grill}}\email{jonas.grill@sap.com}
\author*[2]{\fnm{Thomas} \sur{Bayer}}\email{thomas.bayer@rwu.de}
\author[1,2]{\fnm{Sören} \sur{Berlinger}}\email{soeren.berlinger@sap.com}

\affil[1]{\orgname{SAP SE, Germany}}
\affil[2]{\orgname{Institute for Digital Transformation, Ravensburg-Weingarten University}}
\affil*{\texttt{orcid: 0009-0007-4373-7933}}

\abstract{Accurate extraction of structured information from Safety Data Sheets (SDS) remains challenging in industrial safety due to heterogeneous document formats and the limitations of traditional rule-based methods. This study benchmarks state-of-the-art Large Language Models (LLMs) for automated SDS data extraction, comparing text-based and multimodal processing pipelines. We systematically evaluate four  models—Gemini~1.5~Pro, GPT-4o, Claude 3.7 Sonnet, and Llama 3.1-70B—across three prompting strategies: zero-shot, few-shot, and chain-of-thought. The evaluation framework assessed accuracy, latency, and cost across more than 50,000 extracted data fields. Results show that text-based extraction consistently outperforms multimodal processing across all metrics. Gemini~1.5~Pro combined with a Chain-of-Thought prompt achieved the highest accuracy (84 \%), outperforming GPT-4o (81 \%) and Claude 3.7 Sonnet (79 \%). However, no model surpassed the 90 \% accuracy threshold commonly required for reliable real-world deployment. These findings indicate that general-purpose LLMs are not yet robust enough for unsupervised industrial use, though performance suggests strong potential with task-specific fine-tuning. Future research should focus on domain-adapted training, model calibration, and the integration of Human-in-the-Loop verification to ensure safety-critical reliability.}




\maketitle
\begin{center}
\textit{This manuscript is currently under review at Applied Intelligence.}
\end{center}
\section{Introduction}\label{sec1}

\begin{figure}[h]
\centering
\includegraphics[width=0.75\linewidth]{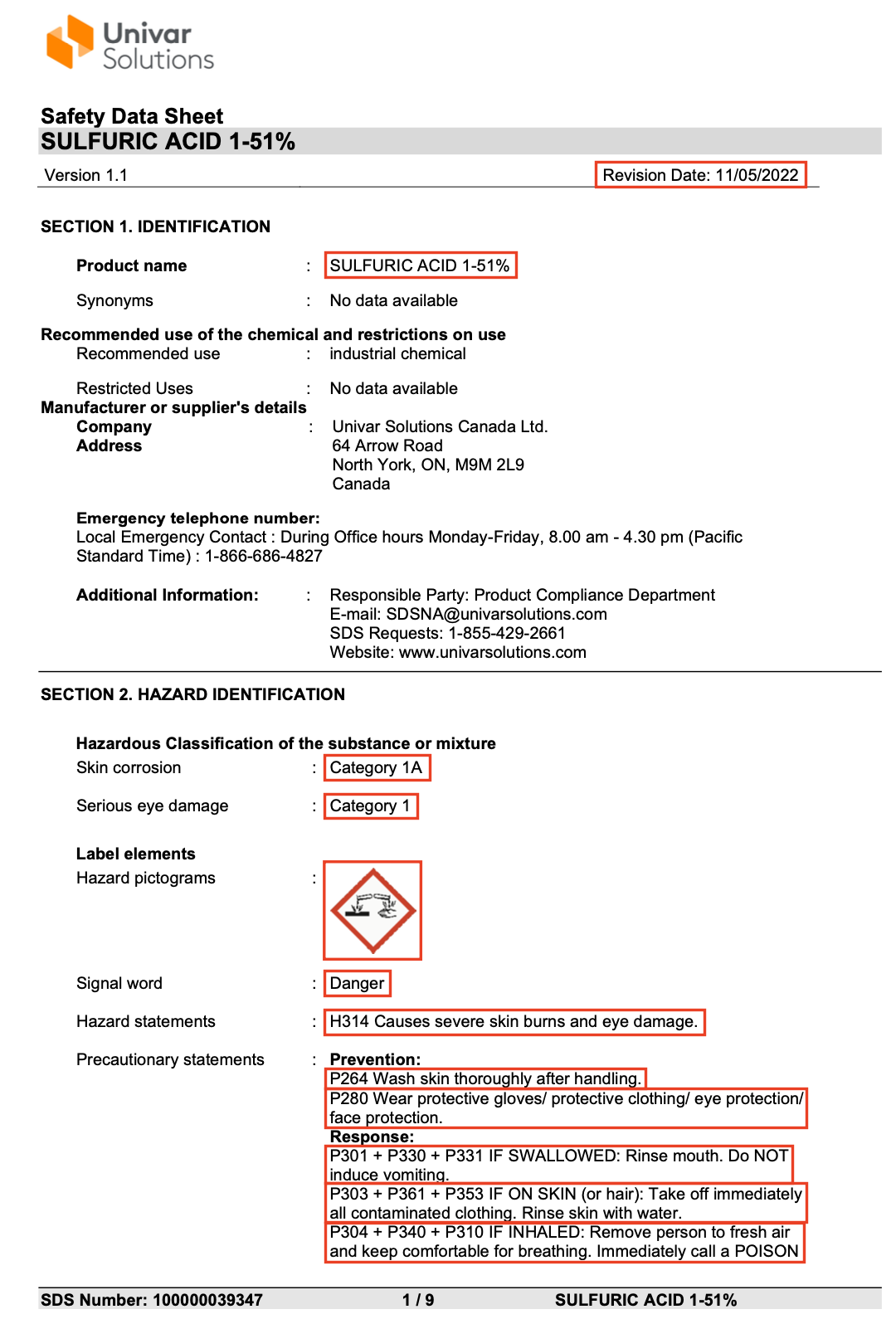}
\caption{Example page from the Safety Data Sheet (SDS) for \sdsexample{} showing annotated regions corresponding to extracted data fields. Source: Univar Solutions USA, Inc.\ \cite{univar2023sds}}
\label{fig:Univar-Solutions-SDS}
\end{figure}

Safety Data Sheets (SDS) are critical regulatory documents in industrial environments, providing authoritative information on hazardous substances, chemical compositions, and protective measures. They form the basis for compliance with international standards such as the Globally Harmonized System (GHS) and the European REACH and CLP regulations \cite{EuropeanCommission2020,OccupationalSafetyHealth2012}. Despite this standardization, SDSs vary considerably in structure, terminology, and completeness across manufacturers. As a result, manual extraction of relevant fields remains costly, time-consuming, and error-prone, limiting scalability and increasing the risk of safety-critical misinterpretation \cite{Khan2025,Matlhare2024}. An example of a SDS is given in Fig. \ref{fig:Univar-Solutions-SDS}.

Recent advances in large language models (LLMs) offer promising opportunities to automate SDS data extraction. Transformer-based LLMs \cite{10.5555/3295222.3295349} exhibit strong semantic and contextual reasoning capabilities and can process semi-structured technical documents beyond the capabilities of traditional rule-based or OCR-centric systems \cite{Zhang2024,Dagdelen2024}. Their in-context learning capabilities allow task specialization without retraining, making them attractive for industrial applications where document heterogeneity is high.

However, their suitability for safety-critical information extraction remains largely unexplored. Existing studies provide limited evidence on how reliably different LLMs extract structured SDS fields, how text-based and multimodal processing compare in performance, and how prompting techniques such as Zero-Shot, Few-Shot, and Chain-of-Thought influence extraction robustness \cite{Opitz2024,Vatsal2024,Sahoo2025,Schulhoff2025,Cheng2025}. A systematic, controlled benchmark addressing these aspects has not yet been conducted.

This paper closes this gap by benchmarking four state-of-the-art LLMs across
text-based and multimodal pipelines: Gemini 1.5 Pro, GPT-4o, Claude 3.7 Sonnet,
and Llama 3.1-70B. We use a schema-constrained protocol to compare accuracy,
latency, and cost across different prompting strategies. We compare accuracy, latency, and token-normalized cost and analyze how prompting influences extraction quality.The remainder is structured as follows: Section 2 outlines foundations; Section 3 details the methodology; Section 4 presents results and discussion; Section 5 concludes with implications and future work.


\section{Related Work}\label{sec2}

The automatic extraction of structured information from Safety Data Sheets (SDS) has been addressed using rule-based, machine learning, and hybrid approaches.
Early large-scale systems rely on hybrid pipelines combining OCR, pattern matching, and neural networks to achieve robust field extraction for regulatory compliance \cite{fenton2021doceng,fenton2023acs, Khan2025}.
Recent machine learning approaches focus on high-precision extraction of standardized SDS fields such as product identifiers, suppliers, and revision dates \cite{khan2025heliyon}, as well as detailed composition data including CAS numbers and concentration values \cite{suman2024scirep}.

With the emergence of large language models (LLMs), generative approaches have been explored for SDS and technical document understanding, showing strong performance in flexible text and table extraction \cite{pekel2025gpt,moreira2024knime}.
Survey work confirms the growing dominance of LLMs for chemical text mining while emphasizing the need for domain adaptation and validation \cite{schilling2025csr}.
In addition, benchmark datasets such as ChemTEB highlight the challenges of SDS-specific language and demonstrate the benefits of domain-specialized embeddings \cite{mansouri2024chemteb}.
Complementary to statistical methods, ontology-based representations using SHACL and SKOS enable semantic validation and integration of extracted SDS data \cite{lu2025shacl}.


\section{Methodology}

We present two variants of an SDS data extraction pipeline—a text-based and an image-based approach—designed to improve the efficiency and accuracy of structured information extraction from PDF documents.
We further introduce a systematic evaluation framework for benchmarking four state-of-the-art \llm s across text-only and multimodal extraction pipelines. The framework explicitly disentangles the effects of model architecture, preprocessing strategy, and prompting methodology on extraction performance, while controlling for dataset composition and output schema consistency. All configurations are evaluated on identical SDS documents using standardized metrics for accuracy, latency, and computational cost.
The following sections describe the methodological design of each approach, highlighting their respective strengths, limitations, and technical trade-offs.

\subsection{Prompt Design}\label{subsec:prompt_design}
Prompt engineering plays a central role in controlling extraction behavior. A unified template combines role specification, task description, schema definition, extraction rules, and strict formatting constraints. We evaluate three prompting strategies, Zero-Shot, Few-Shot, Chain-of-Thought), cf. Table \ref{tab:prompting_strategies}. This setup isolates how prompting techniques influence accuracy, false positives, and extraction stability across models and modalities.

\begin{table}[h]
\centering
\caption{Overview of prompting strategies evaluated in the extraction pipeline.}
\label{tab:prompting_strategies}
\begin{tabular}{p{3cm} p{5cm} p{4cm}}
\hline
\textbf{Strategy} & \textbf{Description} & \textbf{Intended Effect} \\
\hline
Zero-shot prompting &
Direct task instruction without providing examples. The model relies solely on the prompt template and task description. &
Assesses intrinsic generalization capability and baseline extraction performance. \\

Few-shot prompting &
Inclusion of representative input--output examples within the prompt to guide structure and content. &
Improves structural consistency and schema adherence across extractions. \\

Chain-of-thought prompting &
Explicit instruction to produce intermediate reasoning steps prior to generating the final JSON output. &
Enhances reasoning transparency and robustness for complex extraction decisions. \\
\hline
\end{tabular}
\end{table}

Each prompt combines a system instruction (task definition, role, output format, error handling) depending on the prompting strategy, and a JSON schema specifying the structured output of the \llm. An example of a Zero-Shot prompt is given in Fig. \ref{fig:zero_shot_prompt}, and a JSON schema for the SDS \sdsexample \ shown in Fig. \ref{fig:Univar-Solutions-SDS},  concerning First Aid measures can be found in Fig. \ref{fig:json_schema}.

\begin{figure}[h]\label{fig:zero_shot_prompt}
\caption{Zero-shot Prompt: The prompt specifies the extraction task, input
format, schema constraints, and output requirements.}
\begin{tcolorbox}[
  fonttitle=\bfseries,
  title=SDS Extraction Prompt,
  colback=lightestgray,
  colframe=darkgray
]
\begin{lstlisting}[
  % backgroundcolor=\color{lightgray},
  basicstyle=\small\ttfamily,
  breaklines=true,
  breakindent=0pt,
  breakatwhitespace=false,
  columns=fullflexible,
  keepspaces=true,
  showstringspaces=false,
  frame=none,
  xleftmargin=0pt,
  numbers=none,
  aboveskip=0pt,
  belowskip=0pt,
  literate=
  {  - }{{\hspace{1.5em}- }}3
  {- }{{\hspace{1.0em}- }}2
  % literate={- }{{\hspace{1.0em}- }}2
]
# Role
You are a highly capable and precise data extraction specialist for Safety Data Sheets (SDS).

# Task
Extract structured data from the provided Safety Data Sheet (SDS), strictly following the JSON schema below.

# Schema
{schema}

# Input
You will receive:
-  A Safety Data Sheet (SDS) document, provided either directly as a PDF file or as plain text.

# Extraction Rules
- Extract ONLY information from the correct SDS section. DO NOT use information from other sections.
- Match each value precisely to the fields defined in the schema.
- If information is explicitly missing:
  - Use "null" for scalar fields (strings, numbers, booleans).
- Convert all Unicode characters to their closest ASCII equivalent 
- Always use double quotes for JSON keys and values.

# Output Format 
- Output must be ONLY the valid JSON object.
- Start your output with {{ and end with }}.
- DO NOT include markdown, backticks, explanations, schema repetition, comments, or additional text.
\end{lstlisting}
\end{tcolorbox}
\end{figure}

\clearpage
\begin{figure}[h]\label{fig:json_schema}
\caption{JSON for First Aid}
\begin{tcolorbox}[
  breakable,
  enhanced,
  colback=lightestgray,
  colframe=darkgray,
  boxrule=0.5pt,
  arc=2mm,
  left=6pt,
  right=6pt,
  top=6pt,
  bottom=6pt,
  fonttitle=\bfseries\small,
  title=Example JSON Schema Structure for SDS \sdsexample.
]
\begin{lstlisting}[style=jsonbox]
{
  "type": "object",
  "properties": {
    "World_First_Aid_Measures": {
      "type": "object",
      "properties": {
        "First_Aid_Measures": {
          "type": "object",
          "properties": {
            "General_Information": {
              "type": "array",
              "items": { "type": "string" }
            },
            ...,
            "Protection_of_First_Aid_Responders": {
              "type": "array",
              "items": { "type": "string" }
            }
          }
        },
        ...,
        "treatments": {
          "type": "array",
          "items": { "type": "string" }
        }
      }
    }
  }
}
\end{lstlisting}
\end{tcolorbox}
\end{figure}

\subsection{Data Processing Pipeline} \label{subsec:data-processing}

The first approach applies a PDF preprocessing pipeline in which PDF documents are transformed into a structured Markdown representation. This intermediate format enables subsequent text-based information extraction using \llm s. The second approach leverages multimodal LLMs, allowing the system to directly process and interpret textual content from the original PDF without prior format conversion.

Both approaches produce structured metadata and a JSON document containing the extracted fields. Their objective is to streamline the extraction workflow while addressing the architectural and operational characteristics of different AI model classes.

The extraction workflow consists of five sequential stages that transform raw SDS documents into structured JSON outputs while maintaining reproducibility across all configurations.

\begin{enumerate}
\item \textbf{Input and Preprocessing:}
For text-based extraction, native PDF text is extracted using \texttt{PyMuPDF4LLM} and converted to Markdown, preserving structural elements (headings, lists, section boundaries). For multimodal extraction, PDFs are passed directly to vision-capable models via provider-specific APIs: as binary files (Claude~3.7~Sonnet), cloud URIs (Gemini~1.5~Pro), or base64-encoded JPEG images (GPT-4o).

\item \textbf{Prompt Generation:}
Prompt engineering constitutes a central control mechanism for guiding extraction behavior. A standardized prompt template is employed across all experimental settings. This template includes role definition, task specification, schema declaration, extraction constraints, and strict output formatting requirements. Details can be found in Section \ref{subsec:prompt_design}.

\item \textbf{Extraction:}
The prompt and preprocessed document are submitted to the \llm{} API. Text-based requests route through SAP AI Core's Orchestration Service; multimodal requests call provider endpoints directly. Processing time is logged for latency measurement.

\item \textbf{Output Handling and Post-processing:}
The \llm{} returns a structured JSON object conforming to the predefined schema together with metadata including input/output token counts and processing timestamps. The response is parsed to extract valid JSON content. Post-processing includes removal of Markdown fences, schema validation, and Unicode normalization. The cleaned output and associated metadata are stored as uniquely identified files to enable subsequent evaluation and cost computation.

\item \textbf{Data Processing:}
The returned JSON undergoes field-by-field comparison against manually validated ground truth. For each field, a binary match indicator (true/false) is computed and aggregated at the section level to calculate accuracy. Results are stored per SDS document, enabling computation of per-document accuracy across all extracted sections. The results are then aggregated across all ten SDS documents to compute the final accuracy for each model--prompt--method configuration. Token metadata are used to calculate per-document costs based on provider-specific pricing. Finally, a normalized cost function combines accuracy (0.7 weight), processing time (0.2 weight), and cost (0.1 weight) into a unified performance score for systematic comparison across all 21 configurations.

\end{enumerate}

The extraction schema covers a wide range of SDS fields across multiple information types, including textual, numeric, tabular, and graphical elements. SDS sections introduce different structural patterns, from simple key–value pairs (e.g., product identifiers or signal word) to nested lists describing chemical compositions, exposure limits, and regulatory classifications (Figure \ref{fig:Univar-Solutions-SDS}).

Many fields are semantically dense or multi-part, such as tabular data that intermixes quantitative values, units, and qualifiers within a single cell. Examples include concentration ranges, exposure thresholds, and transport information tables that encode several regulatory systems simultaneously. These tabular structures required consistent normalization during parsing to preserve relational meaning across columns. Other elements, such as GHS hazard pictograms, present an additional layer of complexity because they are rendered as images rather than text.

Beyond these special cases, many SDS fields exhibit free-text variability long procedural instructions, embedded lists, and incomplete sentences, which challenge deterministic schema alignment. Certain sections, such as “First Aid Measures” or “Handling and Storage,” frequently contain semi-structured instructions with conditional clauses and implicit references. To handle this, the extraction framework enforced strict JSON conformity while allowing for natural-language variation in values.

Overall, the benchmark encompasses 235 fields, covering all major SDS sections while explicitly including structured, tabular, and visual information. This heterogeneity ensures that the benchmark reflects realistic industrial complexity and tests model robustness across multiple data modalities.

\subsection{Evaluation}

We evaluate on a curated set of ten SDS, comprising approximately 50{,}000 labeled fields. The documents originate from the public ChemicalSafety.com database and represent a diverse range of manufacturers, including BASF, Sigma-Aldrich, and Merck. Only recent revisions were selected to ensure alignment with current GHS and REACH/CLP standards. Each SDS includes ten of the sixteen standardized sections commonly used in industrial workflows. Figure~\ref{fig:Univar-Solutions-SDS} illustrates a representative SDS page with annotated extraction regions. An example of the extracted SDS fields from the SDS \sdsexample, from Fig. \ref{fig:Univar-Solutions-SDS} according to the JSON schema outlined in Fig. \ref{fig:json_schema} is provided in Fig. \ref{fig:json_first_aid_measures}.
\clearpage
\begin{figure}[h]\label{fig:json_first_aid_measures}
\caption{Structured Output for SDS \sdsexample \ of Fig. \ref{fig:Univar-Solutions-SDS}}
\begin{tcolorbox}[
  breakable,
  enhanced,
  colback=lightestgray,
  colframe=darkgray,
  boxrule=0.5pt,
  arc=2mm,
  left=6pt,
  right=6pt,
  top=6pt,
  bottom=6pt,
  fonttitle=\bfseries\small,
  title=Example JSON Schema Structure
]
\begin{lstlisting}[style=jsonbox]
{
  "World_First_Aid_Measures": {
    "First_Aid_Measures": {
      "General_Information": [
        "Take off contaminated clothing."
      ],
      "Following_Inhalation": [
        "Provide fresh air.",
        "In all cases of doubt, or when symptoms persist, seek medical advice."
      ],
      "Following_Skin_Contact": [
        "Rinse skin with water/shower."
      ],
      "Following_Eye_Contact": [
        "Irrigate copiously with clean, fresh water for at least 10 minutes, holding the eyelids apart.",
        "In case of eye irritation consult an ophthalmologist."
      ],
      "Following_Ingestion": [
        "Rinse mouth.",
        "Call a doctor if you feel unwell."
      ],
      "Protection_of_First_Aid_Responders": []
    },
    "Symptoms": [
      "Irritation",
      "Nausea",
      "Vomiting",
      "Gastrointestinal complaints",
      "Headache",
      "Vertigo",
      "Dizziness",
      "Drowsiness",
      "Narcosis"
    ],
    "Treatment": [
      "none"
    ]
  }
}

\end{lstlisting}
\end{tcolorbox}
\end{figure}

Evaluation is based on quality of extraction of SDS fields, the processing time and the number of tokens of the corresponding \llm. The evaluation methodology is provided below.

\begin{enumerate}
\item \textbf{Extraction Quality:} Here, $TP$ and $TN$ denote correct extractions or correct omissions, while $FP$ and $FN$ represent hallucinated or missing fields. A high FP rate indicates hallucinations, which are particularly critical in safety-relevant domains. To capture extraction errors more granularly, we additionally report three complementary quality metrics: Not-Found Rate, False-Positive Rate, and BERTScore for semantic similarity between extracted and reference text \cite{bert-score}. These metrics provide a more differentiated view of omission errors, hallucinations, and semantic similarity.
\begin{itemize}
  \item \textbf{Accuracy} is the proportion of correctly extracted fields relative to all expected fields:
    \[
      \mathrm{Accuracy} =
      \frac{TP + TN}{TP + FP + FN + TN}.
    \]
  \item \textbf{Not-Found Rate (NF Rate):}
    The proportion of required fields that were present in the SDS but not extracted by the model:
    \[
      \mathrm{NF\ Rate} = \frac{FN}{TP + FN}.
    \]
  \item \textbf{False-Positive Rate (FP Rate):}
    The proportion of fields that were extracted despite not being present in the ground truth:
    \[
      \mathrm{FP\ Rate} = \frac{FP}{FP + TN}.
    \]
  \item \textbf{BERTScore (Semantic Similarity):}
    Measures semantic similarity between extracted and reference text by comparing contextual token embeddings using cosine similarity, capturing meaning-level agreement beyond exact string matching \cite{bert-score}.
\end{itemize}

\item \textbf{Processing Time (Latency).}
The end-to-end runtime per document (in seconds), measured from prompt submission to receipt of the final model output.

\item \textbf{Cost (Token Usage).}
Computed from token counts and model pricing:
\[
  \mathrm{Cost} =
  \frac{Tokens_{in}}{10^6} \cdot Price_{in}
  \;+\;
  \frac{Tokens_{out}}{10^6} \cdot Price_{out}.
\]
\end{enumerate}

For a unified comparison across configurations, we report a normalized \textbf{weighted performance score} that emphasizes accuracy (0.7), latency (0.2), and cost (0.1):
\[
\mathrm{Score}
=
0.7 \cdot Accuracy_{\text{norm}}
+ 0.2 \cdot Time_{\text{norm}}
+ 0.1 \cdot Cost_{\text{norm}}.
\]
Normalization is performed across all configurations using min–max scaling.

\begin{table}[!htbp]
\centering
\small
\begin{tabularx}{\columnwidth}{l c X}
\toprule
\textbf{Factor} & \textbf{Weight ($w_i$)} & \textbf{Description} \\
\midrule
Accuracy & 0.7 & Correctness and completeness of the extracted data fields. \\[6pt]
Processing Time & 0.2 & Total time required from start to completion of the extraction process. \\[6pt]
Cost (Tokens) & 0.1 & Computational cost incurred based on the number of tokens processed by the model during extraction. \\
\bottomrule
\end{tabularx}
\caption{Weighted factors used for model evaluation.}
\label{tab:weights}
\end{table}

\subsection{Experimental Procedure}

Four state-of-the-art \llm s were selected to represent major providers:
\textbf{Gemini~1.5~Pro} \textit{(Google DeepMind)}, \textbf{GPT-4o} (\textit{OpenAI}),
\textbf{Claude3.7Sonnet} (\textit{Anthropic}),
and \textbf{Llama~3.1--70B} (\textit{Meta})

Gemini and Claude support native multimodal input; GPT-4o processes image input via base64 encoding; Llama~3.1--70B is text-only and open-source. This selection enables cross-vendor, cross-architecture, and cross-modality comparison under identical extraction conditions.
All model–prompt–method combinations were executed on the full set of ten SDS documents. Each SDS section was processed independently to avoid cross-section leakage and to ensure consistent adherence to the defined JSON schema. Model outputs were compared against a manually validated ground truth, with all reference labels independently verified for consistency and correctness.

For every configuration, field-level metrics—accuracy, false-positive rate, and not-found rate—were computed per section and then averaged across documents. Processing time and token usage were logged for each run to enable cost and latency comparisons. In total, 21 configurations were evaluated under identical conditions to ensure reproducibility and controlled comparison.

\section{Results and Discussion}

\subsection{Comparison of Text-Based and Multimodal Approaches}
\begin{figure}[!ht]
\centering
\includegraphics[width=\linewidth]{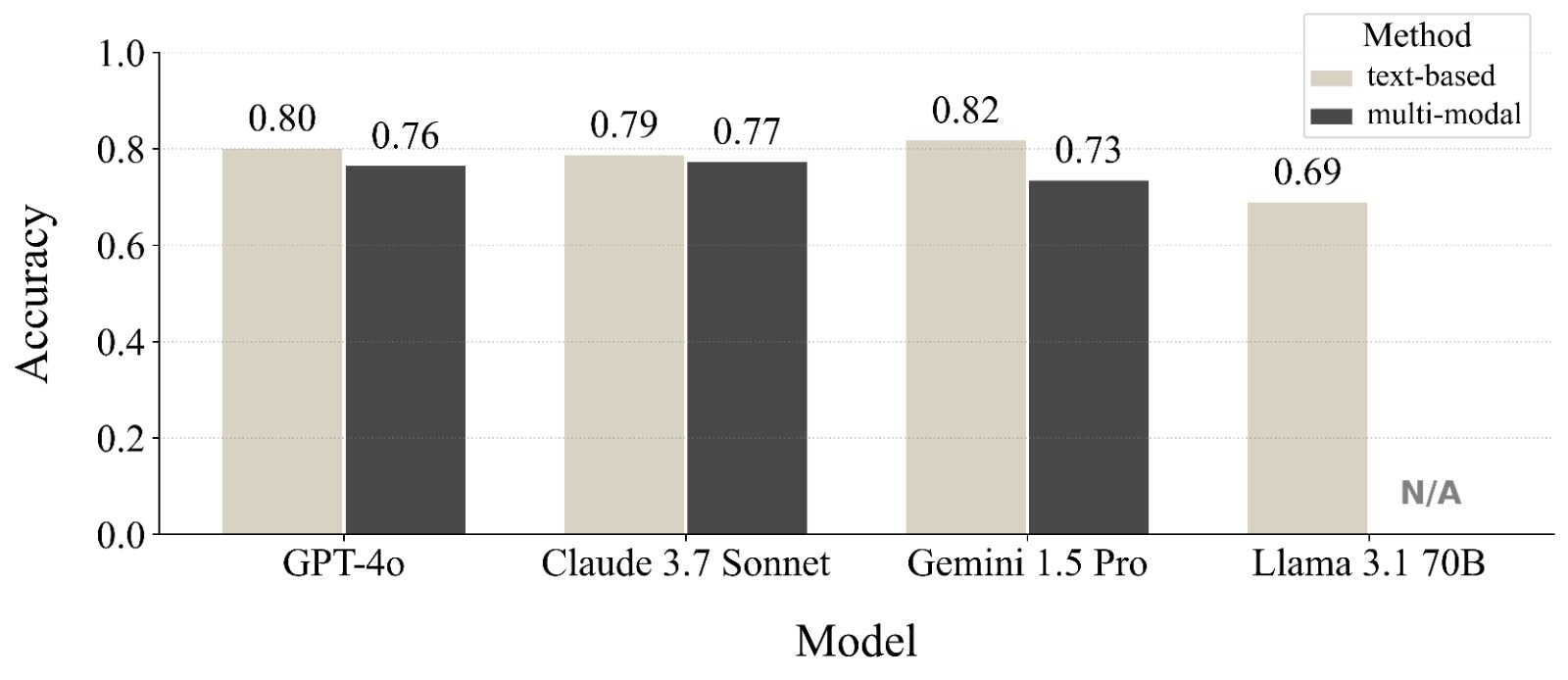}
\caption{Accuracy comparison between text-based and multimodal extraction across all models}
\label{fig:accuracy_models}
\end{figure}

The first part of the analysis compared the text-based and multimodal extraction pipelines across all models.
Results indicate that the text-based approach consistently outperformed the multimodal method in every metric.
While multimodal models can process image-based PDFs directly, their optical character recognition (OCR) introduces additional uncertainty and latency.
Across models, text-based extraction achieved between four and nine percentage points higher accuracy.
As shown in Figure~\ref{fig:accuracy_models}, Gemini~1.5~Pro reached an accuracy of 82~percent in the text-based setup compared to 73~percent in multimodal mode.
Similarly, GPT-4o achieved 80~percent text-based accuracy versus 76~percent in multimodal processing.

\begin{figure}[!ht]
\centering
\includegraphics[width=\linewidth]{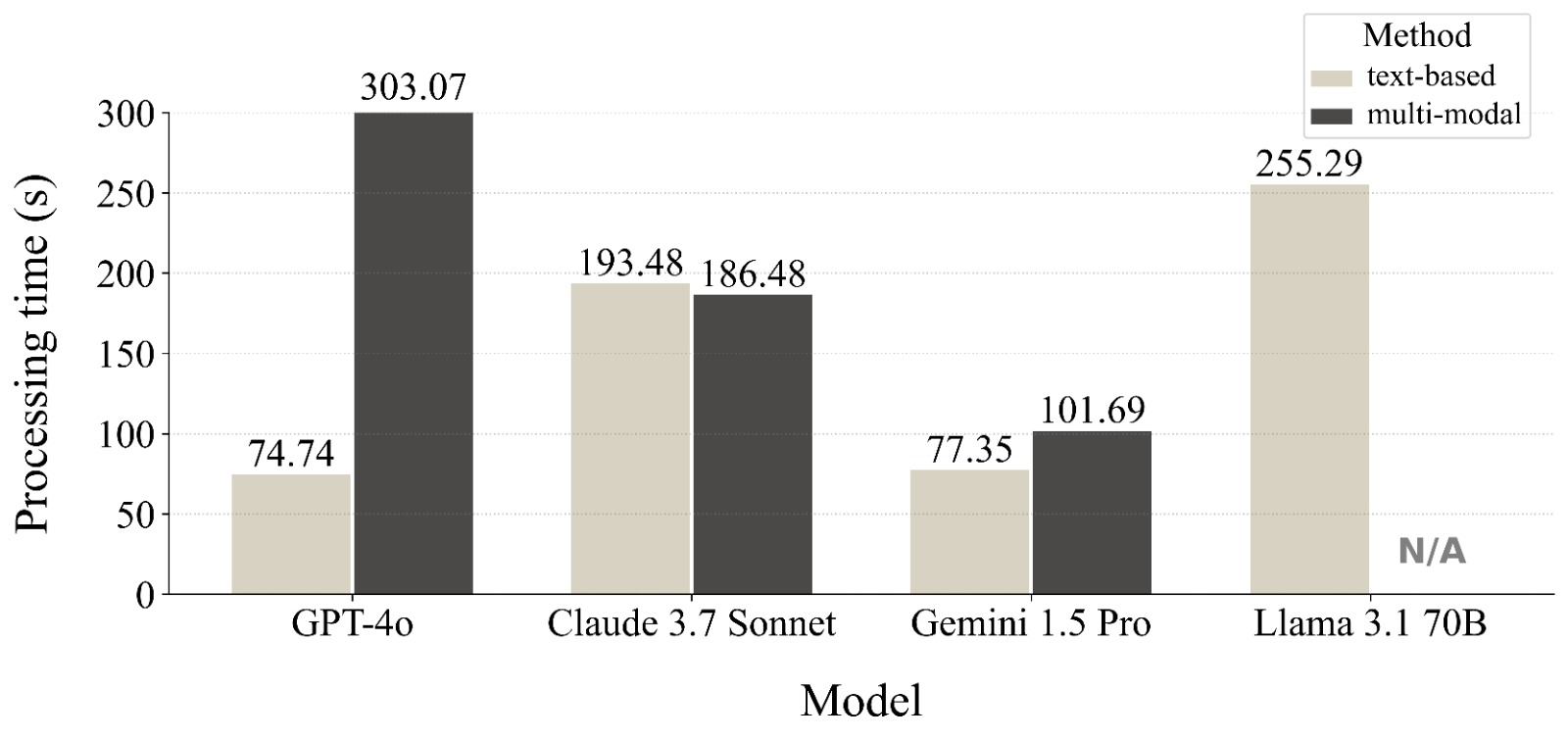}
\caption{Processing time comparison between text-based and multimodal extraction across all models}
\label{fig:processing_time}
\end{figure}

Average processing times were also lower, with GPT-4o completing text-based extractions in approximately 73~seconds compared to nearly 300~seconds in multimodal mode (Figure~\ref{fig:processing_time}).
These results confirm that, for digitally available SDS documents, text-based processing offers a more efficient and reliable approach.

\subsection{Model Performance}
A comparative evaluation of the four Large Language Models (Table~\ref{tab:overall_results_full}) reveals clear performance differences.
\textbf{Gemini~1.5~Pro} achieves the highest overall accuracy in the benchmark (up to 0.84 with Chain-of-Thought prompting) while maintaining one of the lowest total costs, resulting in the most favorable overall efficiency profile.
\textbf{GPT-4o} ranks second, reaching accuracies up to 0.81 and achieving the lowest average processing time (73\,s) among all models, albeit at higher token-related cost.
\textbf{Claude~3.7~Sonnet} performs slightly below GPT-4o with accuracies around 0.79, offering competitive extraction quality but at moderate latency and cost.

In contrast, the open-source \textbf{Llama~3.1--70B} model shows substantially lower accuracy (0.66--0.71 across prompting techniques) and consistently higher false-positive and not-found rates.
The performance gap between the strongest and weakest model configurations spans approximately 18 percentage points, confirming that model choice is the dominant factor affecting SDS extraction quality.

\begin{table*}[!t]
\centering
\caption{Overall evaluation results across all model--prompt--method combinations}
\label{tab:overall_results_full}
\resizebox{\textwidth}{!}{
\begin{tabular}{lllrllllrrrrr}
  \toprule
  Method & Prompt Technique & Model & Accuracy & NF Rate & FP Rate & BERTScore & Time (s) & InToks & OutToks & InCost & OutCost & Total Cost (\$) \\
  \midrule
  text-based & chain-of-thought & gemini-1.5-pro               & \textbf{0.84} & \textbf{0.14} & \textbf{0.11} & \textbf{0.93} & 77.58 & 102518 & 4775 & 0.13 & 0.02 & 0.15 \\
  text-based & chain-of-thought & claude-3.7-sonnet            & 0.79 & \textbf{0.14} & 0.19 & 0.92 & 180.29 & 111253 & 5928 & 0.33 & 0.09 & 0.42 \\
  text-based & chain-of-thought & gpt-4o                       & 0.79 & 0.16 & 0.18 & \textbf{0.93} & 74.03 & 94799  & 4777 & 0.47 & 0.10 & 0.57 \\
  text-based & chain-of-thought & llama-3.1-70b-instruct       & 0.66 & 0.25 & 0.28 & 0.82 & 243.36 & 97603  & 4644 & \textbf{0.07} & \textbf{0.00} & \textbf{0.07} \\
  text-based & zero-shot        & gemini-1.5-pro               & 0.82 & 0.15 & 0.13 & \textbf{0.93} & 76.85 & 99133  & 4948 & 0.12 & 0.02 & 0.15 \\
  text-based & zero-shot        & claude-3.7-sonnet            & 0.78 & 0.15 & 0.20 & 0.91 & 205.96 & 110147 & 6093 & 0.33 & 0.09 & 0.42 \\
  text-based & zero-shot        & gpt-4o                       & 0.81 & \textbf{0.14} & 0.17 & 0.92 & \textbf{73.51} & 97454  & 4846 & 0.49 & 0.10 & 0.58 \\
  text-based & zero-shot        & llama-3.1-70b-instruct       & 0.70 & 0.23 & 0.26 & 0.90 & 250.70 & 90953  & 4768 & \textbf{0.07} & \textbf{0.00} & 0.07 \\
  text-based & few-shot         & gemini-1.5-pro               & 0.79 & 0.18 & 0.17 & 0.92 & 77.62 & 124474 & 4845 & 0.16 & 0.02 & 0.18 \\
  text-based & few-shot         & claude-3.7-sonnet            & 0.79 & \textbf{0.14} & 0.20 & 0.88 & 194.18 & 135777 & 6161 & 0.41 & 0.09 & 0.50 \\
  text-based & few-shot         & gpt-4o                       & 0.80 & 0.15 & 0.17 & 0.91 & 76.69 & 123023 & 5257 & 0.62 & 0.11 & 0.72 \\
  text-based & few-shot         & llama-3.1-70b-instruct       & 0.71 & 0.22 & 0.26 & 0.86 & 271.82 & 121969 & 5235 & 0.09 & \textbf{0.00} & 0.09 \\
  multimodal & chain-of-thought & gemini-1.5-pro               & 0.73 & 0.22 & 0.22 & 0.85 & 100.97 & 57617  & 4237 & 0.07 & 0.02 & 0.09 \\
  multimodal & chain-of-thought & claude-3.7-sonnet            & 0.77 & 0.16 & 0.20 & 0.89 & 178.82 & 129736 & 5665 & 0.39 & 0.08 & 0.47 \\
  multimodal & chain-of-thought & gpt-4o                       & 0.76 & 0.18 & 0.21 & 0.87 & 307.86 & 167783 & 4887 & 0.84 & 0.10 & 0.94 \\
  multimodal & zero-shot        & gemini-1.5-pro               & 0.73 & 0.22 & 0.22 & 0.88 & 101.19 & \textbf{55450}  & \textbf{4187} & \textbf{0.07} & 0.02 & 0.09 \\
  multimodal & zero-shot        & claude-3.7-sonnet            & 0.76 & 0.16 & 0.22 & 0.88 & 184.87 & 112046 & 5741 & 0.34 & 0.09 & 0.42 \\
  multimodal & zero-shot        & gpt-4o                       & 0.76 & 0.18 & 0.22 & 0.88 & 301.51 & 143734 & 5029 & 0.72 & 0.10 & 0.82 \\
  multimodal & few-shot         & gemini-1.5-pro               & 0.74 & 0.20 & 0.21 & 0.86 & 102.92 & 84664  & 4574 & 0.11 & 0.02 & 0.13 \\
  multimodal & few-shot         & claude-3.7-sonnet            & 0.78 & \textbf{0.14} & 0.20 & 0.89 & 195.75 & 134245 & 5784 & 0.40 & 0.09 & 0.49 \\
  multimodal & few-shot         & gpt-4o                       & 0.77 & 0.16 & 0.18 & 0.88 & 299.84 & 159570 & 4909 & 0.80 & 0.10 & 0.90 \\
  \bottomrule
\end{tabular}
}
\end{table*}

\begin{table*}[!t]
\centering
\caption{Accuracy by model, method, and prompting technique}
\label{tab:accuracy}
\resizebox{\textwidth}{!}{%
\begin{tabular}{l|cccc|cccc}
  \toprule
  \textbf{Model} & \multicolumn{4}{c|}{\textbf{Text-Based}} & \multicolumn{4}{c}{\textbf{Multimodal}} \\
  & Chain-of-Thought & Few-Shot & Zero-Shot & Overall & Chain-of-Thought & Few-Shot & Zero-Shot & Overall \\
  \midrule
  Gemini~1.5~Pro    & \textbf{0.84} & 0.79 & 0.82 & 0.82 & 0.73 & 0.74 & 0.73 & 0.73 \\
  Claude 3.7 Sonnet & 0.79 & 0.79 & 0.78 & 0.79 & 0.77 & 0.78 & 0.76 & 0.77 \\
  Llama 3.1 70B     & 0.66 & 0.71 & 0.70 & 0.69 & --   & --   & --   & --   \\
  GPT-4o            & 0.79 & 0.80 & 0.81 & 0.80 & 0.76 & 0.77 & 0.76 & 0.76 \\
  \bottomrule
\end{tabular}
}
\end{table*}

\subsection{Aggregated Cost-Function Evaluation}
The weighted cost-function, combining accuracy (0.7), processing time (0.2), and cost (0.1), provides a holistic efficiency metric.
The highest overall score (0.88) was achieved by Gemini~1.5~Pro using the Chain-of-Thought technique, followed by GPT-4o (0.81) and Claude~3.7~Sonnet (0.72).
Llama~3.1--70B achieved the lowest composite score (0.62).
Contrary to initial assumptions, model efficiency did not strictly correlate with model size or multimodal capabilities; instead, optimized text-based inference produced the most balanced performance.
These outcomes emphasize that text-based extraction, combined with well-structured prompting, currently represents the most viable strategy for SDS information processing.

\begin{figure}[!ht]
\centering
\includegraphics[width=\linewidth]{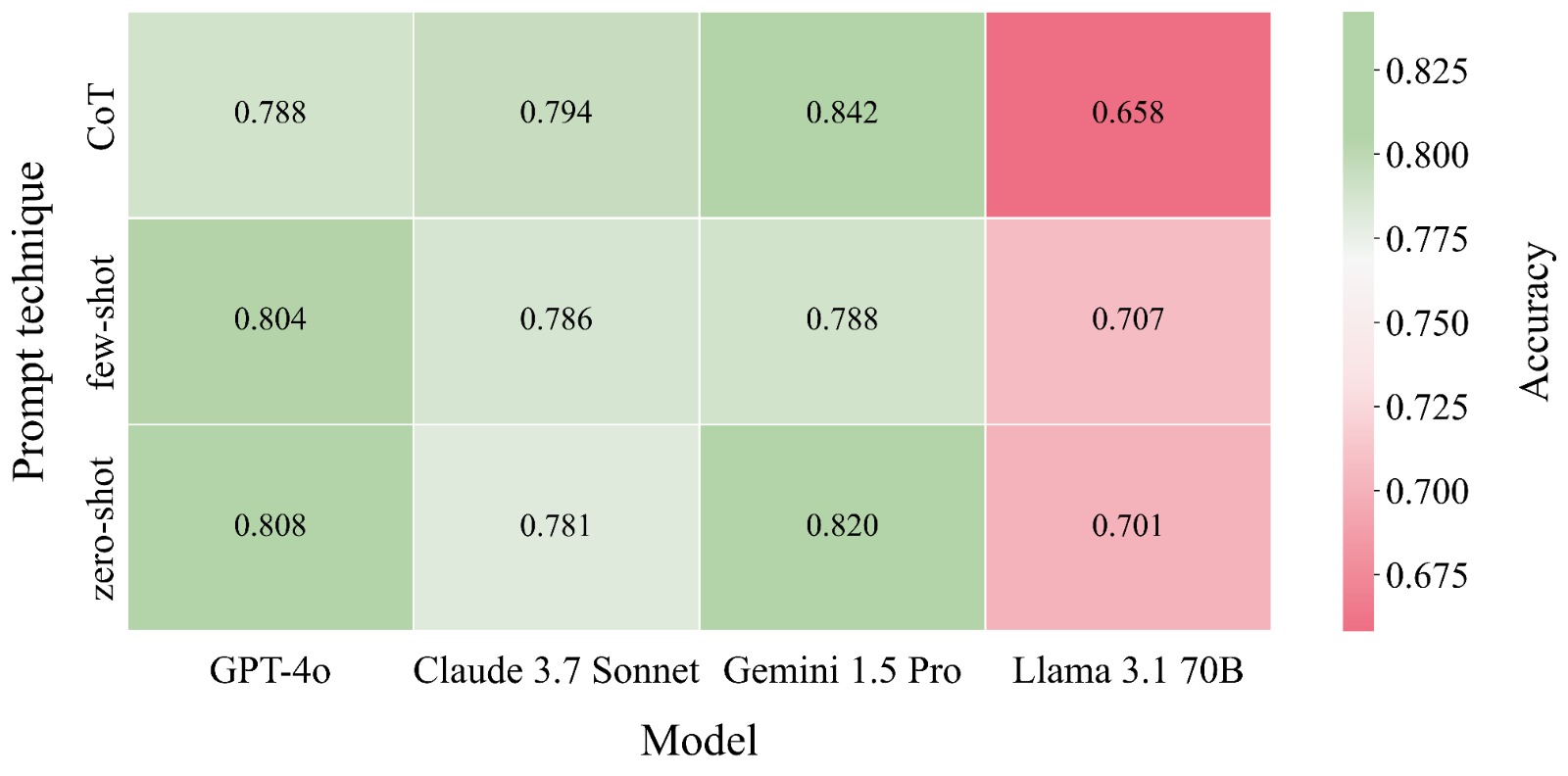}
\caption{Model accuracy by prompting technique (Zero-Shot, Few-Shot, Chain-of-Thought)}
\label{fig:prompt_accuracy}
\end{figure}

\noindent\textit{Prompting effects.}
Across models, Zero-Shot matched or exceeded Few-Shot accuracy while using fewer tokens and incurring lower latency.
This pattern is consistent with the \emph{lost in the middle} effect \cite{liu2023lost}, where models attend less reliably to information positioned mid-prompt; in our setup, adding exemplars increased cost and runtime without improving extraction quality.
Chain-of-Thought helped for Gemini~1.5~Pro but did not close the gap to the 90\% reliability threshold.

\subsection{Aggregated Cost-Function Evaluation}
\begin{figure}[H]
\centering
\includegraphics[width=\linewidth]{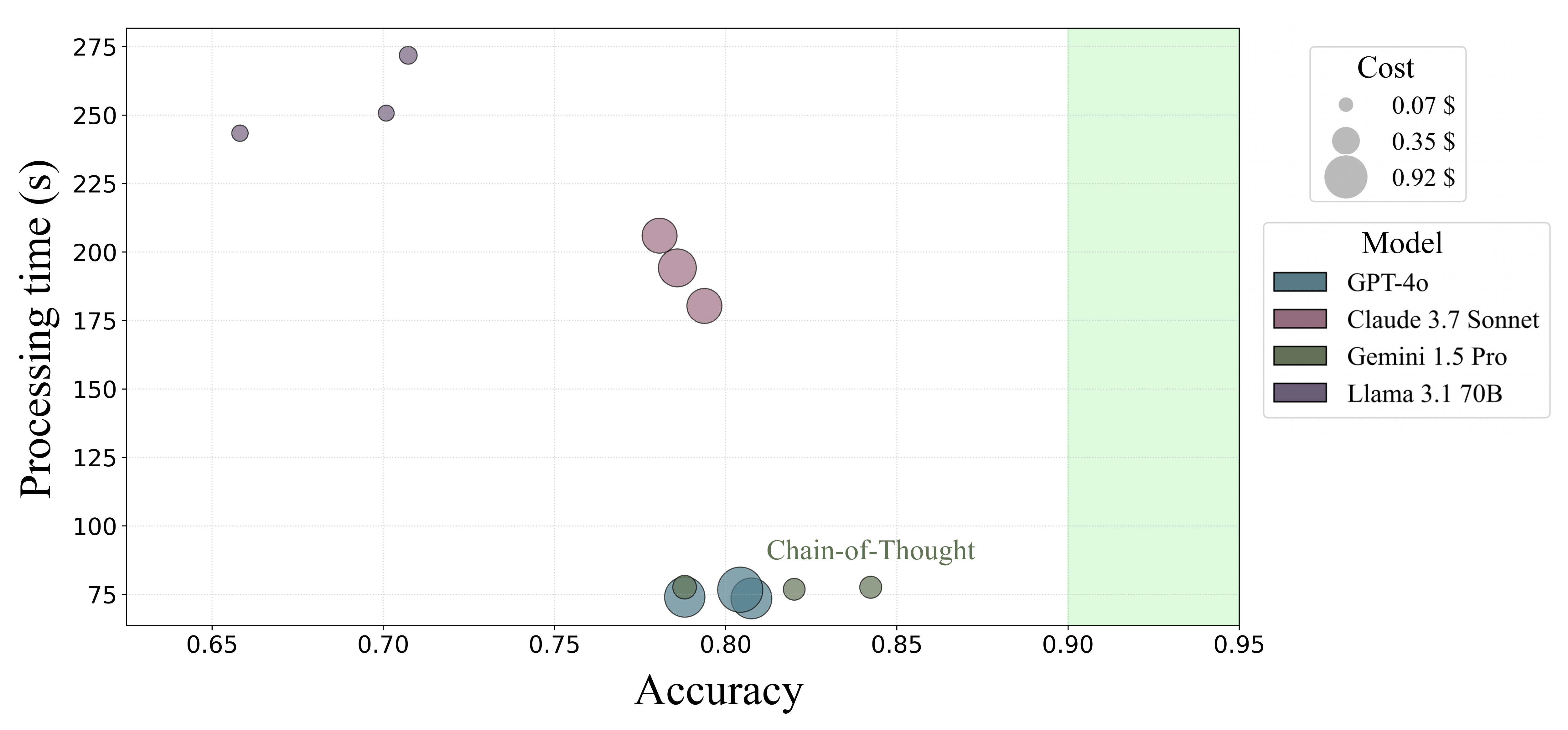}
\caption{Accuracy–processing time–cost trade-off across all model–prompt configurations.
Bubble size encodes total cost; the green band indicates the desired accuracy region ($\geq 0.90$)}
\label{fig:efficiency_tradeoff}
\end{figure}

The weighted cost-function, combining accuracy (0.7), processing time (0.2), and cost (0.1), provides a holistic efficiency metric.
The highest overall score (0.88) was achieved by Gemini~1.5~Pro using the Chain-of-Thought technique, followed by GPT-4o (0.81) and Claude~3.7~Sonnet (0.72).
Figure~\ref{fig:efficiency_tradeoff} visualizes the trade-off between accuracy, latency, and cost across all model--prompt configurations and highlights the relative efficiency of the evaluated models.

Llama~3.1--70B achieved the lowest composite score (0.62).
Contrary to initial assumptions, model efficiency did not strictly correlate with model size or multimodal capabilities; instead, optimized text-based inference produced the most balanced performance.
Taken together, these results indicate that text-based extraction, combined with well-structured prompting, currently represents the most viable strategy for SDS information processing.

\subsection{Discussion of Findings}
The experimental results demonstrate that current state-of-the-art LLMs can extract structured information from SDS documents with promising but still limited accuracy.
Although the best configuration achieved 84~percent accuracy, no model surpassed the 90~percent reliability threshold required for autonomous deployment in industrial safety applications.
High false-positive rates, particularly among smaller or multimodal models, remain a key limitation.
Nevertheless, the observed performance suggests that domain-specific fine-tuning or hybrid workflows incorporating Human-in-the-Loop validation could bridge this gap.
Moreover, continuous benchmarking across model versions will be essential to track future progress as foundation models evolve rapidly.
Overall, this study confirms the potential of LLMs for semi-automated safety data extraction while underscoring the need for further optimization and rigorous validation before operational integration.

\paragraph{Limitations and Responsible Use.}
Limitations include residual false positives that matter in safety-critical settings, sub-90\% accuracy across all configurations, OCR-induced artifacts in multimodal runs, and uncertain generalizability beyond the ten SDS. To mitigate risk, we recommend a human-in-the-loop validator for high-impact fields (e.g., hazards, first aid, PPE) and audit logging for traceability. Future work should expand the dataset, report inter-annotator agreement, stress-test degraded scans, and evaluate domain-adapted fine-tuning and calibration for reliable confidence estimates.

\section{Conclusion and Future Work}

The benchmark of four state-of-the-art Large Language Models—Gemini~1.5~Pro, GPT-4o, Claude~3.7~Sonnet, and Llama~3.1–70B—showed that text-based processing consistently outperforms multimodal approaches across accuracy, runtime, and cost. The best configuration, Gemini~1.5~Pro with Chain-of-Thought prompting, achieved 84\% accuracy; however, no model reached the 90\% reliability threshold required for autonomous use in safety-critical settings. False positives remain the most consequential failure mode, and multimodal OCR artefacts further limit extraction robustness. Model choice exerted a far stronger influence on performance than prompting strategies, indicating that architectural capabilities dominate over interaction design. Given these limitations and the narrow scope of the ten-document dataset, practical deployments should incorporate a Human-in-the-Loop validation step to mitigate risks and ensure the trustworthy handling of chemical safety information. The presented framework provides a reproducible basis for future evaluation, fine-tuning, and optimization of LLM-based extraction pipelines in regulated industrial environments.

\section*{Declarations}

\subsection*{Conflict of Interest}
The authors declare that they have no competing interests. The first and third authors are affiliated with SAP; these affiliations did not influence the design, results, or interpretation of the research.

\subsection*{Data availability}

The datasets generated and/or analysed during the current study are available from the corresponding author on reasonable request.
The code used for the analysis is available from the corresponding author upon reasonable request.




\bibliography{benchmarking_llm_SDS_Grill}

\end{document}